\documentclass[11pt]{article}

\usepackage[margin=1in]{geometry}

\usepackage{times}
\usepackage[T1]{fontenc}
\usepackage{float}

\usepackage{amsmath}
\usepackage{amssymb}

\usepackage[numbers,sort&compress]{natbib}

\usepackage{amsmath,amsfonts,bm}

\def\eqref#1{equation~\ref{#1}}

\def\1{\bm{1}}

\DeclareMathAlphabet{\mathsfit}{\encodingdefault}{\sfdefault}{m}{sl}
\SetMathAlphabet{\mathsfit}{bold}{\encodingdefault}{\sfdefault}{bx}{n}

\usepackage{hyperref}
\usepackage{url}

\usepackage{algorithm}
\usepackage{algpseudocode}

\usepackage{graphicx}
\usepackage{tabularx}
\usepackage{multirow}
\usepackage{subcaption}

\usepackage{tikz}
\usepackage{circuitikz}
\usepackage{pgfplots}
\pgfplotsset{compat=1.18}

\title{CMDO: A Cognitive Memory-Driven Optimization Algorithm for Adaptive Population-Based Search}

\author{
Mohammed Yusuf Mujawar\thanks{These authors contributed equally to this work.}, 
Shahram Rahimi, 
Noorbakhsh Amiri Golilarz\footnotemark[1]\\
Department of Computer Science\\
The University of Alabama, AL, USA\\
}

\date{}

\begin{document}

\maketitle

\begin{abstract}
Population-based optimization methods often use previous search information
through successful solutions, parameter adaptation, or operator performance,
but they rarely retain the context in which a search behavior succeeded or
failed. We introduce Cognitive Memory-Driven Optimization (CMDO), a
derivative-free population-based optimizer that represents experience as the
relationship between search context, search behavior, and observed outcome.
CMDO organizes these experiences across working, episodic, and consolidated
memory, retrieves them according to similarity with the current search state,
and uses both positive and negative evidence to guide subsequent search.
Retrieved experience does not replay previous candidate locations; instead, it
selects search recipes that are reconstructed from the current population
through exploratory, directed, and local search behaviors with adaptive search
geometry. We evaluate CMDO on selected Blackbox Optimization Benchmarking test suite on COCO (BBOB/COCO) and Congress on Evolutionary Computation 2017 (CEC2017) problems against
DE, CMA-ES, SHADE, GWO, HHO, and ORCA, and further study its application to
seven-parameter photovoltaic model estimation using measured current--voltage
data. The results show problem-dependent but competitive optimization
performance, including the lowest median error among the compared methods on
CEC2017 F10. More importantly, analysis of the search traces shows that
context-dependent recall changes the distribution of executed search
behaviors, while unsuccessful experiences remain available as negative
evidence for later decisions, showing that accumulated experience directly
influences subsequent search behavior. These results support the use of explicit
context--behavior--outcome memory as an active mechanism for controlling
population-based search.
\end{abstract}


\section{Introduction}

Optimization methods must continually balance exploration of new regions with refinement of promising solutions. Existing approaches achieve this through population dynamics, stochastic search, parameter adaptation, or information retained from previous evaluations. Differential Evolution and CMA-ES adapt search through population differences and evolving sampling distributions \citep{storn1997differential,hansen2003cmaes}, while methods such as Grey Wolf Optimization (GWO), Harris Hawks Optimization (HHO), and ORCA optimization use alternative population-based search dynamics \citep{mirjalili2014gwo,heidari2019hho,golilarz2020orca}. Other approaches explicitly use search history; for example, SHADE adapts parameters from successful past values, and adaptive operator-selection methods use previous operator performance to guide future choices \citep{tanabe2013shade,durgut2021aos}. These methods show the value of past information, but they also raise a more fundamental question: \emph{what should an optimizer remember?}

A successful search step is meaningful not only because it improved a solution, but also because of the conditions under which it worked. The same behavior may be useful in one stage of optimization and ineffective in another. This suggests that optimization memory should capture more than a good solution, parameter value, or operator score. Instead, it can represent the relationship between the \emph{current search state}, the \emph{behavior applied in that state}, and the \emph{outcome that followed}. Such a representation allows previous experience to be reused only when it is relevant to the present search condition.

This idea is closely related to the functional role of memory in neurocognitive systems, where memory supports adaptation through context-sensitive storage, retrieval, and reuse of experience \citep{golilarz2026neurocognitive}. Related work in episodic control, memory-augmented neural models, and Hebbian memory has similarly shown the value of recalling prior experience according to the current context \citep{blundell2016modelfree,pritzel2017neural,rae2020compressive,wu2022memorizing,money2026hebbian,mujawar2026adaptivehebbian,mujawar2026hierarchicalhebbian}. These ideas motivate a shift from treating memory as passive storage toward using it as an active mechanism for controlling future behavior.

We introduce \emph{Cognitive Memory-Driven Optimization} (CMDO), an optimization framework in which search experience is used to guide future decisions. CMDO stores experiences that connect search context, search behavior, and observed outcome. These experiences are maintained across working, episodic, and consolidated memory, retrieved according to contextual similarity, and updated through reinforcement, consolidation, and forgetting. CMDO also retains negative experience, allowing previously unsuccessful behavior to reduce the likelihood of repeating similar search decisions under comparable conditions. Retrieved experience then influences how exploratory, directed, and local search behaviors are constructed. 

We evaluate CMDO on the selected problems from BBOB/COCO benchmark suite
\citep{hansen2021coco} and the CEC2017 single-objective
bound-constrained benchmark suite \citep{awad2016cec2017}, and on
photovoltaic parameter estimation using a two-diode model and measured
current--voltage data \citet{muhammad2019pv}. Comparisons include DE, CMA-ES, SHADE, GWO, HHO, and ORCA under matched experimental conditions. The experiments assess both optimization performance and the contribution of the proposed memory mechanisms.

The main contributions of this work are:
\begin{itemize}
    \item We introduce CMDO, a memory-driven optimization framework that represents experience through the relationship between search context, applied behavior, and observed outcome.
    \item We develop a multi-timescale memory architecture with context-dependent retrieval, reinforcement, consolidation, forgetting, and continued acquisition of experience.
    \item We incorporate both positive and negative search experience so that useful behaviors can be reinforced while previously unsuccessful behaviors are discouraged in related search states.
    \item We evaluate CMDO across numerical benchmarks and photovoltaic parameter estimation.
\end{itemize}

The remainder of this paper is organized as follows.
Section~2 reviews related work on adaptive optimization, episodic and
context-dependent memory, and neurocognitive memory mechanisms.
Section~3 presents CMDO, including its search context, search recipes,
multi-timescale memory, contextual retrieval, and memory update process.
Section~4 reports the numerical benchmark and photovoltaic parameter-estimation
results and analyzes how recalled experience changes the search behavior, and discusses the implications and limitations of the proposed approach. Section~5 concludes the paper. Additional algorithmic, implementation, and
experimental details are provided in the appendix.

\section{Related Work and Motivation}

\subsection{Memory and adaptation in optimization}

Optimization algorithms use past search information in different ways. Differential Evolution and CMA-ES adapt search through population relationships and evolving sampling distributions \citep{storn1997differential,hansen2003cmaes}, while methods such as GWO, HHO, and ORCA define alternative population-based search dynamics \citep{mirjalili2014gwo,heidari2019hho,golilarz2020orca}. More explicit use of search history appears in adaptive evolutionary methods. SHADE stores successful control-parameter values and uses them to guide future parameter generation \citep{tanabe2013shade}, while adaptive operator-selection methods update operator preferences according to observed performance \citep{durgut2021aos}. Related studies further show that adaptive parameter control can strongly influence Differential Evolution behavior \citep{tanabe2020landscapes,tanabe2020parametercontrol}.

These approaches demonstrate the value of past performance, but CMDO focuses on a different question: \emph{under what search condition did a particular behavior succeed or fail?} The usefulness of a search action may depend on population diversity, search progress, stagnation, and proximity to promising regions. CMDO therefore links search behavior to the state in which it was applied, rather than treating success as independent of context.

\subsection{Episodic and context-dependent memory}

Context-dependent reuse of experience has been studied extensively outside numerical optimization. Model-Free Episodic Control and Neural Episodic Control store prior experiences and retrieve related states to support later decisions \citep{blundell2016modelfree,pritzel2017neural}. Experience replay similarly reuses previous interactions during learning, with prioritized replay emphasizing experiences expected to provide stronger learning signals \citep{schaul2016prioritized}. Generalizable episodic memory and episodic curiosity further show how stored experience can be compared with current representations to influence future behavior \citep{hu2021generalizable,savinov2019episodic}. Memory-augmented sequence models provide a complementary perspective. Compressive Transformers, Memorizing Transformers, and Recurrent Memory Transformers maintain information beyond the immediate processing window \citep{rae2020compressive,wu2022memorizing,bulatov2022recurrent}. Although these methods address different tasks, they support a common principle relevant to CMDO: memory is most useful when the system determines what should be retained, when it should be retrieved, and how it should affect current computation.

\subsection{From neurocognitive memory to search experience}

Neurocognitive-inspired intelligence treats memory as an active component of adaptation rather than passive storage. The framework in \citet{golilarz2026neurocognitive} describes working memory as a rapidly accessible store for current information and longer-term memory as a mechanism for retaining context-rich and consolidated experience. Retrieval is associative and sensitive to context, while consolidation allows selected experience to persist and guide future behavior. Recent Hebbian memory models provide related computational examples. Hebbian fast weights form temporary associative memories during an episode \citep{money2026hebbian}, adaptive Hebbian routing regulates memory contribution, plasticity, and retention according to the current task \citep{mujawar2026adaptivehebbian}, and Hierarchical Hebbian Memory organizes experience across working, episodic, and more stable memory levels \citep{mujawar2026hierarchicalhebbian}.

CMDO transfers these functional ideas to optimization, but uses a different memory object. Rather than storing only neural representations, candidate solutions, successful parameters, or aggregate operator rewards, CMDO stores a \emph{search experience}, the search state, the behavior applied in that state, and the resulting outcome. These experiences are maintained across working, episodic, and consolidated memory and are retrieved according to similarity with the current search context. CMDO also retains negative evidence. A behavior that failed under a similar search condition can reduce support for repeating that behavior, while successful experience can strengthen it. Thus, memory is used not only to recall what worked, but also to avoid repeating previously unproductive search behavior.

The resulting distinction is central to CMDO: rather than adapting parameters independently, selecting operators from aggregate reward, or recalling earlier solutions, CMDO retrieves context-conditioned experience describing \emph{what was tried, under what search condition, and what consequence followed}. The next section formalizes how these experiences are represented, stored, retrieved, and converted into new optimization actions.

\section{Cognitive Memory-Driven Optimization Algorithm}
\label{sec:cmdo}

\subsection{Overview}

We consider bounded continuous optimization,
\begin{equation}
    \min_{\mathbf{x}\in\Omega} f(\mathbf{x}),
\end{equation}
where $\Omega \subset \mathbb{R}^{D}$ is the search space. CMDO maintains a
population of candidate solutions and generates one new candidate at each
search step.

\begin{figure}[t]
    \centering
    \includegraphics[
        width=\linewidth,
        height=0.22\textheight
    ]{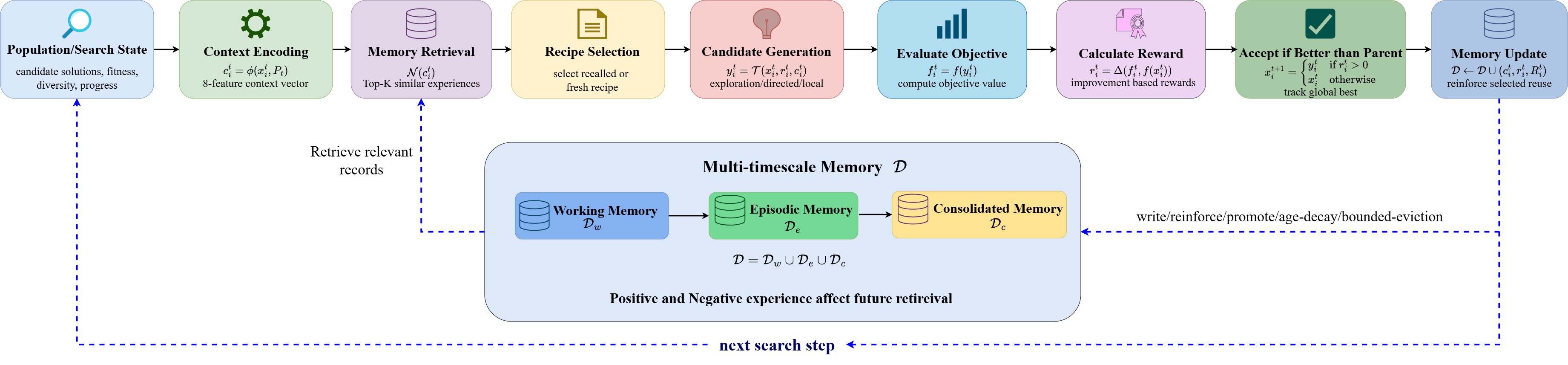}
    \caption{
    Overall architecture of Cognitive Memory-Driven Optimization (CMDO).
    }
    \label{fig:cmdo_architecture}
\end{figure}

The main difference in CMDO is how previous search information is represented.
Instead of storing only a good solution or a successful parameter value, CMDO
stores a complete search experience,
\begin{equation}
    \mathcal{E}
    =
    \left(
        \mathbf{c},
        \mathbf{r},
        R
    \right),
    \label{eq:experience}
\end{equation}
where $\mathbf{c}$ is the current search context, $\mathbf{r}$ is the search
recipe used in that context, and $R$ is the resulting reward. When a similar
search condition appears later, CMDO retrieves relevant past experiences and
uses them to guide the next search action. The overall CMDO architecture is illustrated in
Figure~\ref{fig:cmdo_architecture}. The search process forms a closed loop in
which the current population determines the search context, memory influences
the next search behavior, and the observed outcome is written back as new
experience for future retrieval.

\subsection{Search context}

The search context summarizes the current condition of the population. CMDO
uses an eight-dimensional context vector,
\begin{equation}
    \mathbf{c}
    =
    [c_1,c_2,\ldots,c_8]
\end{equation}

The eight components describe: (1)~population spread, (2)~the rank of the focal solution, (3)~its distance to the current best solution, (4)~its distance to the population centroid, (5)~the fraction of the evaluation budget consumed, (6)~current stagnation, (7)~the recent frequency of positive rewards, and (8)~the fitness gap relative to the current best solution. These features allow CMDO to distinguish between different search situations.
The context uses only information available during optimization; knowledge of
the true optimum is never used by the search controller.

\subsection{Search recipe and candidate generation}

A search recipe defines how a new candidate should be generated:
\begin{equation}
    \mathbf{r}
    =
    (o,s,m,g),
\end{equation}
where $o$ is the search operator, $s$ is the step scale, $m$ controls the
directional mixture, and $g$ determines whether the search acts on the full
space or on a subset of dimensions. CMDO uses three search behaviors which are, \emph{exploration}, \emph{directed search},
and \emph{local refinement}. Let $\mathbf{x}_i$ be the focal solution,
$\mathbf{x}_{best}$ the current best solution, and
$\boldsymbol{\delta}$ a difference vector obtained from two population
members. Let $\boldsymbol{\epsilon}$ denote a population-scaled random
perturbation. The three search directions are
\begin{align}
    \mathbf{d}_{exp}
    &=
    m\boldsymbol{\delta}
    +
    (1-m)\boldsymbol{\epsilon},
    \\
    \mathbf{d}_{dir}
    &=
    m(\mathbf{x}_{best}-\mathbf{x}_i)
    +
    (1-m)\boldsymbol{\delta},
    \\
    \mathbf{d}_{loc}
    &=
    0.1
    \left[
        m(\mathbf{x}_{best}-\mathbf{x}_i)
        +
        (1-m)\boldsymbol{\epsilon}
    \right]
\end{align}

A new candidate is generated as
\begin{equation}
    \mathbf{x}'
    =
    \mathbf{x}_i
    +
    s\mathbf{d}
    \label{eq:candidate}
\end{equation}

The recipe therefore controls both the type and strength of the search move.
The geometry component can restrict the move to a randomly selected subset of
approximately $\sqrt{D}$ dimensions, allowing CMDO to alternate between
full-space and lower-dimensional search.

\subsection{Multi-timescale memory}

CMDO organizes search experience into three bounded memory levels,
\emph{working memory}, \emph{episodic memory}, and \emph{consolidated memory}.
Working memory stores recent experiences, episodic memory retains a larger
history over a longer period, and consolidated memory preserves patterns that
have repeatedly produced useful outcomes. The three levels store the context, recipe, and outcome associated with search
behavior rather than only candidate locations. This allows CMDO to recall
\emph{how to search} rather than simply \emph{where a good solution was found}.

\subsection{Context-dependent retrieval}

For the current context $\mathbf{c}$ and a stored context $\mathbf{c}_j$, CMDO
measures their similarity using
\begin{equation}
    S_j
    =
    \exp
    \left(
        -
        \frac{
            \operatorname{MSE}
            (\mathbf{c},\mathbf{c}_j)
        }{
            2h^2
        }
    \right),
    \label{eq:similarity}
\end{equation}
where $h$ controls how strongly context differences affect retrieval.

Older memories gradually lose influence through a retention factor,
\begin{equation}
    W_j
    =
    S_j \rho^{a_j},
\end{equation}
where $a_j$ is the age of the memory and $\rho$ is the retention coefficient. CMDO keeps only sufficiently relevant memories and considers the most relevant
records for recipe selection. A stored experience receives reuse credit only
when its recipe is actually selected and executed.

\begin{table}[t]
\centering
\small
\renewcommand{\arraystretch}{1.25}
\setlength{\tabcolsep}{2.5pt} 

\caption{Median final error over three seeds on BBOB and CEC2017 benchmarks (lower is better).}
\label{tab:benchmark_results}
\vspace{4pt}

\begin{tabularx}{\linewidth}{|l|
    >{\raggedleft\arraybackslash\hsize=1.20\hsize}X|
    >{\raggedleft\arraybackslash\hsize=0.85\hsize}X|
    >{\raggedleft\arraybackslash\hsize=0.95\hsize}X|
    >{\raggedleft\arraybackslash\hsize=1.30\hsize}X|
    >{\raggedleft\arraybackslash\hsize=0.85\hsize}X|
    >{\raggedleft\arraybackslash\hsize=0.85\hsize}X|}
\hline
\multirow{2}{*}{\textbf{Method}} & \multicolumn{3}{c|}{\textbf{BBOB ($D=5$)}} & \multicolumn{3}{c|}{\textbf{CEC2017 ($D=10$)}} \\ \cline{2-7}
 & \textbf{F1} & \textbf{F8} & \textbf{F15} & \textbf{F1} & \textbf{F4} & \textbf{F10} \\ \hline
CMDO   & $1.450\times 10^{-3}$ &  5.7813 & 21.4916 & $3.7848\times 10^{9}$  &  189.9318 & 1322.7172 \\ \hline
HHO    & $2.406\times 10^{-1}$ & 35.9889 & 47.1264 & $1.1906\times 10^{10}$ &  566.7792 & 1622.6580 \\ \hline
GWO    & $1.088\times 10^{-2}$ &  3.9398 & 21.7654 & $3.8409\times 10^{8}$  &   15.6910 & 1415.7922 \\ \hline
ORCA   & $1.038\times 10^{-2}$ &  3.2394 & 11.5687 & $1.2090\times 10^{10}$ & 1900.3090 & 1586.3345 \\ \hline
DE     & $5.859\times 10^{-4}$ &  3.8711 & 20.5967 & $9.6074\times 10^{6}$  &    7.6000 & 1946.1414 \\ \hline
CMA-ES & $9.861\times 10^{-3}$ &  4.5501 & 31.7369 & $1.7650\times 10^{7}$  &    8.4741 & 2070.8946 \\ \hline
SHADE  & $1.263\times 10^{-1}$ & 21.2723 & 15.0919 & $3.3633\times 10^{7}$  &   16.4747 & 1800.6883 \\ \hline
\end{tabularx}
\end{table}

\subsection{Learning from positive and negative experience}

CMDO learns from both successful (positive) and unsuccessful (negative) search behavior. Positive
experience supports repeating a recipe when a related context appears again,
while negative experience reduces support for similar behaviors that
previously failed under comparable conditions.

The support for a retrieved recipe is
\begin{equation}
    Q_j
    =
    \text{positive support}
    -
    \lambda\,
    \text{negative evidence},
    \label{eq:support}
\end{equation}
where $\lambda$ controls the influence of failure information.

Negative evidence is applied only when the failed experience is behaviorally
similar to the candidate recipe, including the same search operator and
geometry. If no recalled recipe has sufficient support, CMDO generates a new
recipe. A fixed probability of generating a fresh recipe is also retained so
that new search behaviors can continue to be discovered.

\subsection{Reward and memory update}

After evaluating a candidate, CMDO compares its objective value with that of
the focal solution. The reward is computed as
\begin{equation}
    R
    =
    \tanh
    \left(
        \frac{
            f(\mathbf{x}_i)-f(\mathbf{x}')
        }{
            \sigma_f
        }
    \right),
    \label{eq:reward}
\end{equation}
where $\sigma_f$ is a robust scale estimate from the current population
fitness values.

A positive reward indicates improvement, while a non-positive reward records
an unsuccessful action. The candidate replaces the focal solution only when
it improves the objective. The resulting experience is stored in working and
episodic memory, and repeatedly useful experience may later be consolidated.
If the executed recipe was recalled, only the selected source receives reuse
feedback.

The overall CMDO cycle can be summarized as

\begin{center}
\small
\text{Observe}
$\rightarrow$
\text{Recall/Explore}
$\rightarrow$
\text{Act}
$\rightarrow$
\text{Evaluate}
$\rightarrow$
\text{Update}
$\rightarrow$
\text{Remember}
\end{center}

The cycle emphasizes that memory controls how search behavior is selected,
while each recalled recipe is reconstructed from the current population rather
than replaying a previously visited solution.

\section{Experimental Results and Discussion}
\label{sec:results}

\subsection{Numerical optimization performance}

Table~\ref{tab:benchmark_results} reports the median final
optimization error over seeds 15--17. Lower values indicate better performance.
The two benchmark suites are presented separately because their objective
scales differ substantially. On BBOB F1, CMDO achieves a median error of $1.45\times10^{-3}$, lower than
HHO, GWO, ORCA, CMA-ES, and SHADE. On F8 it improves over HHO and SHADE,
while on F15 it remains close to GWO and DE and improves over HHO and
CMA-ES. The results indicate that the fixed CMDO configuration remains
competitive on the selected BBOB problems, although its relative
performance varies by problem.

On CEC2017 F1 and F4, CMDO improves over HHO and ORCA, while several
evolutionary baselines obtain lower errors. On F10, CMDO reaches a median
error of $1322.72$, lower than all six comparison methods. This is the
strongest numerical benchmark result for CMDO in the reported experiments. Figure~\ref{fig:cec_f10_convergence} shows that CMDO continues to improve
throughout the available evaluation budget on F10 and finishes with a lower
median best-so-far error than HHO, GWO, and ORCA.

\subsection{How memory changes the search}

The main question behind CMDO is whether stored experience actually changes
future search behavior. We examine this by separating newly generated
\emph{fresh} recipes from recipes selected through memory recall.

\begin{figure}[t]
\centering

\begin{subfigure}[t]{0.485\linewidth}
    \centering
    \includegraphics[width=\linewidth, height=4.6cm]{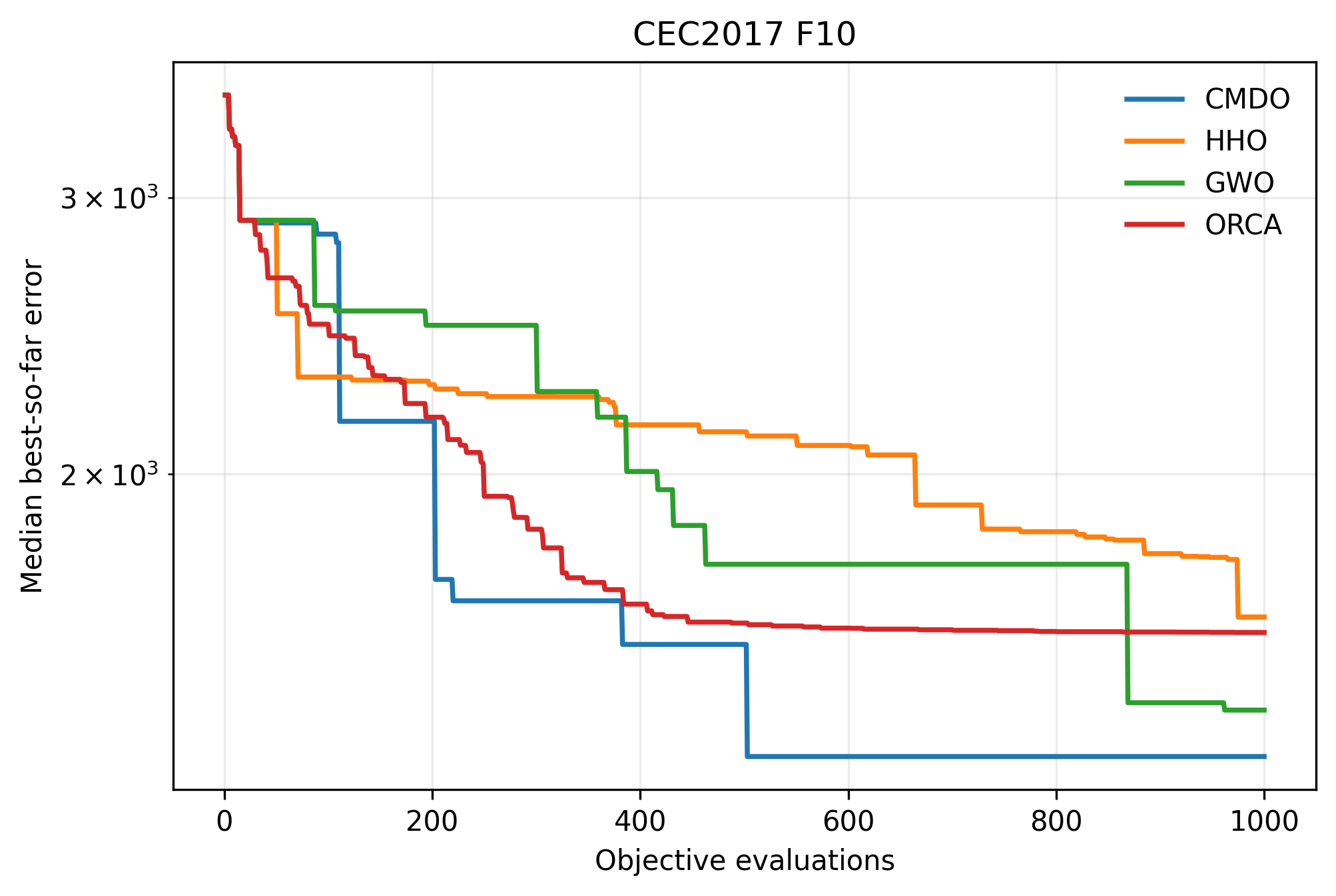}
    \caption{Convergence on CEC2017 F10 (median best-so-far error over seeds 15--17; lower is better).}
    \label{fig:cec_f10_convergence}
\end{subfigure}
\hfill
\begin{subfigure}[t]{0.485\linewidth}
    \centering
    \includegraphics[width=\linewidth, height=4.6cm]{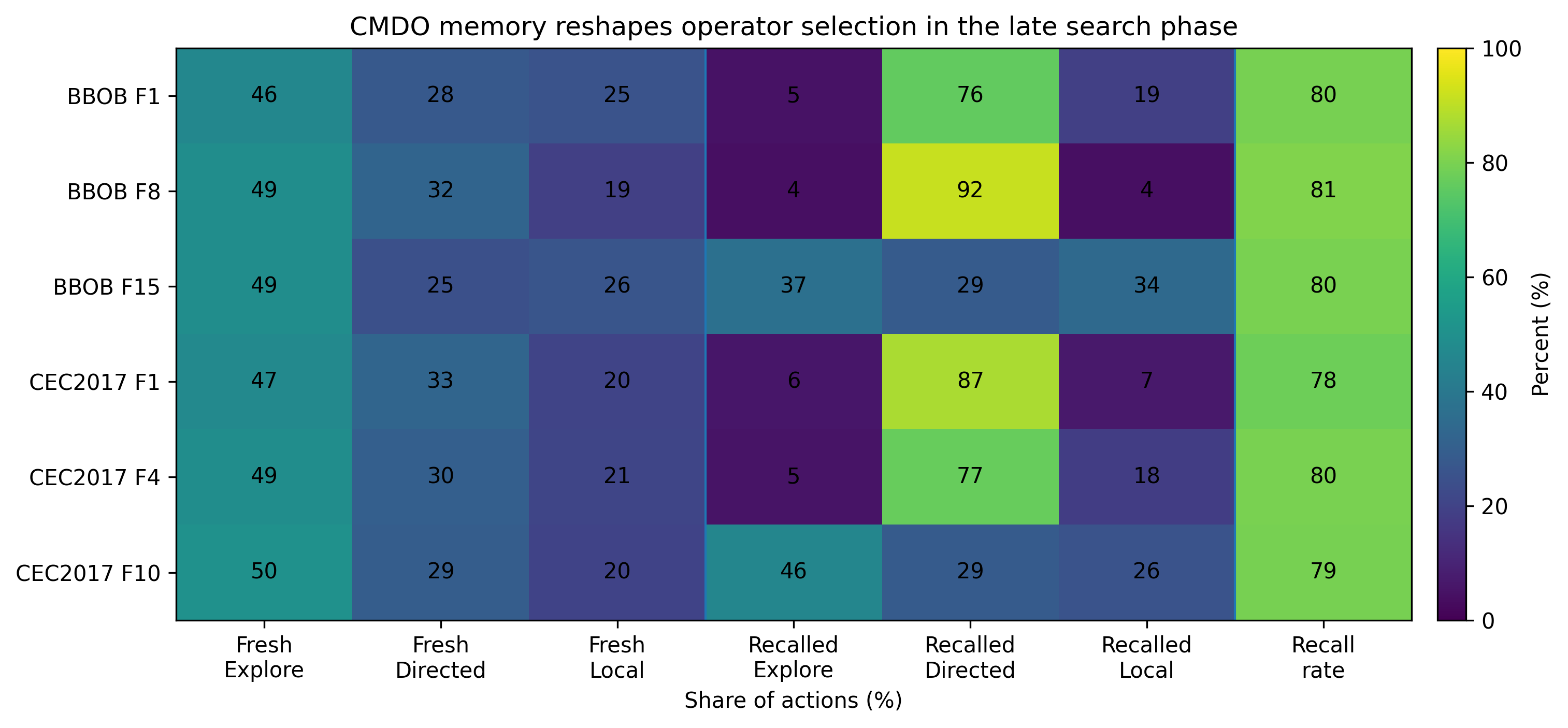}
    \caption{Memory-driven search behavior in the late optimization phase.}
    \label{fig:memory_behavior}
\end{subfigure}

\caption{Comparison of convergence and memory-driven search behavior.}
\label{fig:overall_behavior}
\end{figure}

Figure~\ref{fig:memory_behavior} shows a clear difference between fresh and
recalled behavior. Fresh recipes remain close to the predefined operator
sampling distribution, whereas recalled recipes develop problem-dependent
preferences. Directed search becomes dominant among recalled actions on
BBOB F1, BBOB F8, CEC2017 F1, and CEC2017 F4, while BBOB F15 shows a more
balanced mixture of exploration, directed search, and local refinement.
CEC2017 F10 exhibits another pattern, with recalled actions maintaining a
mixed operator distribution. Because the fresh-recipe distribution is fixed, these shifts arise from
accumulated search experience rather than from a change in the underlying
operator prior. Memory therefore does more than store previous evaluations, it changes which
search behaviors are selected under related search conditions. The variation
across benchmark functions further indicates that CMDO does not converge to a
single globally preferred operator.

\subsection{Negative experience and memory persistence}

CMDO retains unsuccessful as well as successful search experiences. Negative
records remain available during retrieval and can reduce support for similar
recipes when related contexts are encountered again. This is particularly
visible on CEC2017 F10, where the late search memory contains substantially
more negative than positive original experiences. These failures are therefore
not discarded simply because they did not improve the focal solution. The three memory levels provide complementary time scales for this experience.
Working memory retains recent events, episodic memory maintains a larger
history of positive and negative outcomes, and consolidated memory preserves
patterns that have demonstrated useful reuse. Together with age-dependent
retention and bounded capacity, this allows recent evidence to remain
responsive while repeatedly useful behavior can persist for longer. These observations support the central design of CMDO in which the optimizer remembers
\emph{what was tried, under what search condition, and what happened}, using
both successful and unsuccessful experience to influence later behavior.

\subsection{Photovoltaic parameter estimation}

We further evaluate CMDO on the RTC France two-diode photovoltaic
parameter-estimation problem using the measured current--voltage data reported
by \citet{muhammad2019pv}. The experiment uses 26 measured $I$--$V$
observations collected at $33^{\circ}\mathrm{C}$ and
$1000~\mathrm{W/m^2}$. All compared methods use the same measured data,
parameter bounds, objective definition, and budget of 2000 objective
evaluations per run. Results are reported over seeds 101--110.

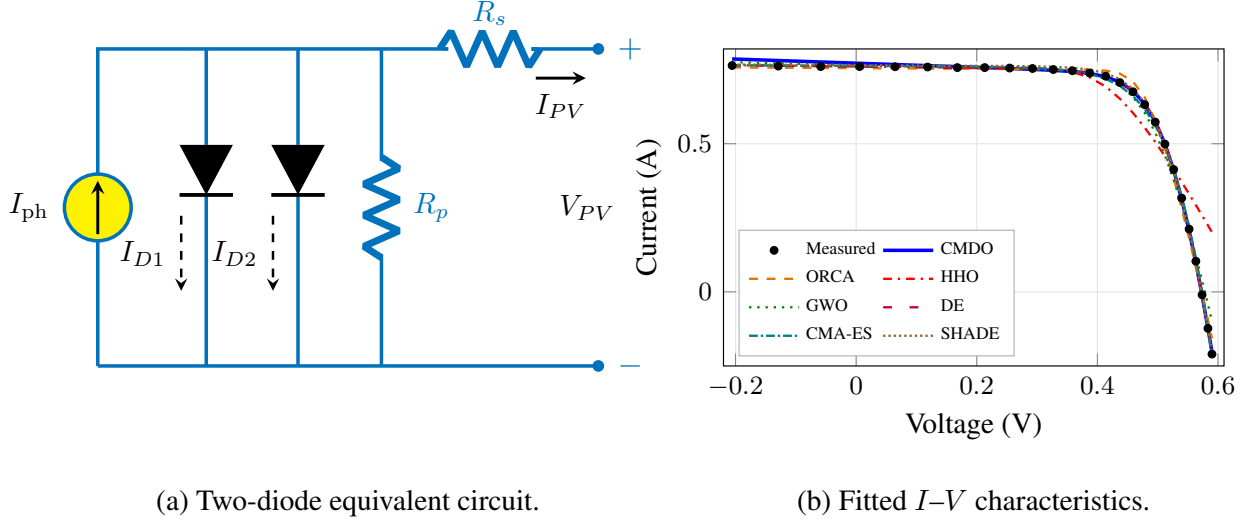
\begin{figure}[t]
\centering

\resizebox{\linewidth}{!}{%
\begin{tikzpicture}

\begin{scope}[shift={(0,0)}]

\definecolor{pvblue}{RGB}{0,114,188}
\definecolor{pvyellow}{RGB}{255,242,0}
\definecolor{pvblack}{RGB}{0,0,0}

\draw[pvblue, line width=1.2pt] (0,0) -- (0,3.8) -- (4.2,3.8);
\draw[pvblue, line width=1.2pt] (0,0) -- (6.0,0);

\draw[pvblue, line width=1.2pt] (4.2,3.8) to[R, color=pvblue] (5.2,3.8) -- (6.0,3.8);
\node[font=\small\bfseries, text=pvblue, anchor=south] at (4.7,3.95) {$R_s$};

\fill[pvyellow] (0,1.9) circle [radius=0.40];
\draw[pvblue, line width=1.2pt] (0,1.9) circle [radius=0.40];
\draw[pvblack, ->, >=stealth, line width=1.0pt] (0,1.58) -- (0,2.22);
\node[font=\small\bfseries, text=pvblack, anchor=east] at (-0.45,1.9) {$I_{\mathrm{ph}}$};

\draw[pvblue, line width=1.2pt] (1.3,3.8) -- (1.3,2.65);
\draw[pvblue, line width=1.2pt] (1.3,2.05) -- (1.3,0);
\fill[pvblack] (0.98,2.65) -- (1.62,2.65) -- (1.3,2.05) -- cycle;
\draw[pvblack, line width=1.3pt] (0.98,2.05) -- (1.62,2.05);

\draw[pvblack, dashed, ->, >=stealth, line width=0.85pt] (1.0,1.8) -- (1.0,0.9);
\node[font=\small\bfseries, text=pvblack, anchor=east] at (0.95,1.35) {$I_{D1}$};

\draw[pvblue, line width=1.2pt] (2.4,3.8) -- (2.4,2.65);
\draw[pvblue, line width=1.2pt] (2.4,2.05) -- (2.4,0);
\fill[pvblack] (2.08,2.65) -- (2.72,2.65) -- (2.4,2.05) -- cycle;
\draw[pvblack, line width=1.3pt] (2.08,2.05) -- (2.72,2.05);

\draw[pvblack, dashed, ->, >=stealth, line width=0.85pt] (2.1,1.8) -- (2.1,0.9);
\node[font=\small\bfseries, text=pvblack, anchor=east] at (2.05,1.35) {$I_{D2}$};

\draw[pvblue, line width=1.2pt] (3.4,3.8) to[R, color=pvblue] (3.4,0);
\node[font=\small\bfseries, text=pvblue, anchor=west] at (3.65,1.9) {$R_p$};

\draw[pvblack, ->, >=stealth, line width=1.0pt] (5.25,3.45) -- (5.85,3.45);
\node[font=\small\bfseries, text=pvblack, anchor=north] at (5.55,3.4) {$I_{PV}$};

\fill[pvblue] (6.0,3.8) circle (0.07);
\fill[pvblue] (6.0,0) circle (0.07);
\node[font=\normalsize\bfseries, text=pvblue, anchor=west] at (6.1,3.8) {$+$};
\node[font=\normalsize\bfseries, text=pvblue, anchor=west] at (6.1,0) {$-$};

\node[font=\small\bfseries, text=pvblack, anchor=west] at (5.4,1.9) {$V_{PV}$};

\node[font=\small, anchor=north] at (3.0, -1.35) {(a) Two-diode equivalent circuit.};

\end{scope}

\begin{scope}[shift={(7.5,0)}]

\begin{axis}[
    width=6.0cm,
    height=3.8cm,
    scale only axis, 
    xlabel={Voltage (V)},
    ylabel={Current (A)},
    xmin=-0.22,
    xmax=0.61,
    ymin=-0.25,
    ymax=0.82,
    grid=major,
    major grid style={gray!20},
    axis line style={black},
    tick label style={font=\footnotesize},
    label style={font=\small},
    xlabel style={yshift=2pt},
    ylabel style={yshift=-2pt},
    legend style={
        at={(0.03,0.03)},
        anchor=south west,
        font=\tiny,
        draw=black!30,
        fill=white,
        cells={anchor=west}
    },
    legend columns=2
]

\addplot[
    only marks,
    mark=*,
    mark size=1.3pt,
    black
]
table[
    x=V,
    y=Measured,
    col sep=comma
]{iv_curves_median_run_overleaf.csv};
\addlegendentry{Measured}

\addplot[very thick, blue]
table[x=V, y=CMDO, col sep=comma]{iv_curves_median_run_overleaf.csv};
\addlegendentry{CMDO}

\addplot[thick, orange!90!black, dashed]
table[x=V, y=ORCA, col sep=comma]{iv_curves_median_run_overleaf.csv};
\addlegendentry{ORCA}

\addplot[thick, red, dashdotted]
table[x=V, y=HHO, col sep=comma]{iv_curves_median_run_overleaf.csv};
\addlegendentry{HHO}

\addplot[thick, green!55!black, dotted]
table[x=V, y=GWO, col sep=comma]{iv_curves_median_run_overleaf.csv};
\addlegendentry{GWO}

\addplot[thick, purple, loosely dashed]
table[x=V, y=DE, col sep=comma]{iv_curves_median_run_overleaf.csv};
\addlegendentry{DE}

\addplot[thick, teal, densely dashdotted]
table[x=V, y=CMAES, col sep=comma]{iv_curves_median_run_overleaf.csv};
\addlegendentry{CMA-ES}

\addplot[thick, brown!80!black, densely dotted]
table[x=V, y=SHADE, col sep=comma]{iv_curves_median_run_overleaf.csv};
\addlegendentry{SHADE}

\end{axis}

\node[font=\small, anchor=north] at (3.0, -1.35) {(b) Fitted $I$--$V$ characteristics.};

\end{scope}

\end{tikzpicture}%
}

\vspace{2pt}
\caption{Photovoltaic parameter-estimation setup and response: (a) equivalent circuit of the seven-parameter two-diode model, and (b) measured versus fitted $I$--$V$ characteristics for the RTC France solar cell at $33^{\circ}\mathrm{C}$ and $1000~\mathrm{W/m^2}$.}
\label{fig:pv_combined}
\end{figure}

Figure~\ref{fig:pv_combined}a illustrates the equivalent circuit used in
the experiment. The seven optimized parameters are
$I_{\mathrm{ph}}$, $I_{01}$, $I_{02}$, $R_s$, $R_p$, $a_1$, and $a_2$.
For each candidate parameter vector, the two-diode model is evaluated at the
measured voltage points and the mean absolute error between measured and
modeled current is used as the optimization objective. In addition, Figure~\ref{fig:pv_combined}b compares the measured RTC France
$I$--$V$ observations with fitted responses produced by CMDO and the six
comparison methods. For each optimizer, the displayed curve corresponds to
the run closest to that method's ten-seed median MAE, providing a
representative visualization of its fitted parameter set. The curves reproduce
the overall nonlinear shape of the measured response to different degrees,
with the most visible deviations occurring around the knee and high-voltage
region of the characteristic.

Table~\ref{tab:pv_results} summarizes the corresponding optimization
performance over the ten runs. CMDO obtains a median MAE of $0.005402$~A,
a mean MAE of $0.010135$~A, and a median RMSE of $0.007720$~A. Its median
MAE is lower than those of HHO, GWO, ORCA, and SHADE, while DE and CMA-ES
obtain lower median values. The fitted responses in Figure~\ref{fig:pv_combined}b complement these
aggregate errors by showing how the estimated model parameters translate into
the resulting current--voltage characteristics.

\begin{table}[t]
\centering
\small
\renewcommand{\arraystretch}{1.25} 
\setlength{\tabcolsep}{4pt}       

\newcolumntype{Y}{>{\raggedleft\arraybackslash}X}

\caption{RTC France two-diode parameter-estimation results over ten seeds. Errors are in amperes.}
\label{tab:pv_results}
\vspace{4pt}

\begin{tabularx}{\linewidth}{|l|Y|Y|Y|>{\centering\arraybackslash}X|}
\hline
\textbf{Method} & \textbf{Median MAE} & \textbf{Mean MAE} & \textbf{Median RMSE} & \textbf{CMDO wins} \\ \hline
CMDO   & $0.005402$ & $0.010135$ & $0.007720$ & --    \\ \hline
HHO    & $0.067527$ & $0.066728$ & $0.106345$ & 10/10 \\ \hline
GWO    & $0.015631$ & $0.017507$ & $0.025084$ & 7/10  \\ \hline
ORCA   & $0.014460$ & $0.019404$ & $0.020608$ & 8/10  \\ \hline
DE     & $0.001841$ & $0.002706$ & $0.002306$ & 2/10  \\ \hline
CMA-ES & $0.005338$ & $0.005368$ & $0.006985$ & 6/10  \\ \hline
SHADE  & $0.005724$ & $0.005646$ & $0.006698$ & 6/10  \\ \hline
\end{tabularx}
\end{table}

The paired-seed comparison provides an additional view of run-to-run behavior.
CMDO obtains a lower MAE than HHO in all ten paired runs, ORCA in eight, GWO
in seven, and both CMA-ES and SHADE in six. Against DE, CMDO obtains the lower
MAE in two of the ten runs. CMA-ES has a slightly lower median MAE, while CMDO
obtains lower MAE in six paired seeds, illustrating why aggregate statistics
and paired comparisons can provide complementary views of optimizer behavior.

The photovoltaic experiment uses the same CMDO memory architecture and search
mechanism as the numerical benchmark experiments. No photovoltaic-specific
operator or memory rule is introduced. The experiment therefore provides a
second application setting in which the context--behavior--outcome memory
representation is used to guide optimization of a nonlinear physical model.

\subsection{Discussion}
\label{sec:discussion}

The results show that CMDO's main contribution lies in how search experience is
used to influence future decisions. Recalled recipes develop different
operator preferences from freshly generated recipes, indicating that memory
actively reshapes the search rather than serving only as passive storage. The
variation across benchmark functions further suggests that this adaptation is
context dependent rather than driven by one globally preferred behavior. The use of both positive and negative experience is important to this process.
Successful behaviors can gain support in related search states, while
unsuccessful behaviors remain available as evidence against repeating similar
decisions. Combined with working, episodic, and consolidated memory, this gives
CMDO a mechanism for balancing recent experience with more persistent search
knowledge. The photovoltaic experiment provides a complementary test outside synthetic
benchmarks. The same CMDO memory architecture is applied to the two-diode
parameter-estimation problem without introducing a domain-specific search rule,
showing that the proposed context--behavior--outcome representation can also
guide nonlinear physical parameter estimation. The present evaluation is limited to a compact set of low-dimensional
benchmarks and one measured photovoltaic curve. Future work should examine
higher-dimensional, noisy, constrained, dynamic, and multi-objective settings,
as well as whether useful experience can be transferred across related
optimization tasks. Overall, the findings support the central idea of CMDO: optimization memory can
be more useful when it captures \emph{what was tried, under what search
condition, and what consequence followed}, allowing past experience to become
an active part of search control.

\section{Conclusion}
\label{sec:conclusion}

We introduced Cognitive Memory-Driven Optimization (CMDO), an optimization
framework that represents search experience through the relationship between
context, behavior, and outcome. Rather than remembering only successful
solutions or parameter values, CMDO retrieves relevant past experience and
uses it to shape new search actions under similar conditions. Across the selected BBOB and CEC2017 problems and photovoltaic parameter-estimation experiments, CMDO
shows competitive problem-dependent performance while the memory analysis
demonstrates that recalled experience changes the distribution of executed
search behaviors. The results also show that both successful and unsuccessful
experience can contribute to later decisions through contextual retrieval,
reinforcement, and consolidation. More broadly, CMDO provides a way to separate the search mechanism itself from
the memory that governs when different behaviors should be reused. This makes
it possible to study optimization not only in terms of which operator performs
well, but also in terms of which experiences remain useful across changing
search conditions. Such a perspective may help support more adaptive
optimizers in which search behavior is shaped by accumulated experience rather
than by fixed heuristics alone. These findings support a broader view of optimization memory as an active
search-control mechanism rather than a passive record of previous evaluations.
Future work will investigate richer context representations, larger and more
diverse optimization settings, and the transfer of useful search experience
across related problems.

\section*{Acknowledgment}

The authors acknowledge the support and resources provided by the
Bioinspired Robotics, AI, Imaging and Neurocognitive Systems (BRAINS)
Laboratory at The University of Alabama.

\bibliography{cmdo_references}
\bibliographystyle{iclr2027_conference}

\appendix
\section*{Appendix}

\appendix

\section{CMDO Algorithmic Details}
\label{app:cmdo_details}

This appendix provides implementation details for CMDO that complement the
higher-level description in Section~\ref{sec:cmdo}. The algorithm operates in
a normalized search space $[0,1]^D$, while objective evaluations are performed
after mapping candidate solutions to the physical problem bounds. All function
evaluations, including initialization, count toward the evaluation budget.

\subsection{Overall CMDO procedure}

Algorithm~\ref{alg:cmdo} summarizes the complete optimization loop. At each
step, one focal population member is selected cyclically. CMDO describes the
current search state, retrieves relevant experience, selects either a recalled
or fresh recipe, constructs one candidate, evaluates it, and records the
resulting experience. A recalled recipe does not replay an old candidate vector. The recipe specifies
a form of search behavior that is reconstructed using the current population.
Consequently, the same remembered recipe can produce different displacements
when it is reused in different search states.

\subsection{Search context}

For focal solution $\mathbf{z}_i$, CMDO uses the eight-dimensional context
\begin{equation}
    \mathbf{c}_i =
    [c_i^{(1)},\ldots,c_i^{(8)}]
\end{equation}

The components represent population spread, focal rank, distance to the current
best, distance to the population centroid, consumed evaluation budget,
stagnation, recent positive-reward frequency, and focal-to-best fitness gap.
The implementation computes them as
\begin{align}
c_i^{(1)}
&=
\min\left(
1,\,
2\,\operatorname{mean}_{d}
[\operatorname{std}(Z_{:,d})]
\right),
\\
c_i^{(2)}
&=
\frac{
\sum_{j=1}^{N}\mathbb{I}[f_j<f_i]
}{
\max(1,N-1)
},
\\
c_i^{(3)}
&=
\frac{\|\mathbf{z}_i-\mathbf{z}_{best}\|_2}{\sqrt{D}},
\\
c_i^{(4)}
&=
\frac{\|\mathbf{z}_i-\bar{\mathbf{z}}\|_2}{\sqrt{D}},
\\
c_i^{(5)}
&=
\frac{n_{\mathrm{eval}}}{B},
\\
c_i^{(6)}
&=
\min\left(
1,\,
\frac{s_{\mathrm{stag}}}{10N}
\right),
\\
c_i^{(7)}
&=
\frac{1}{|H|}
\sum_{R\in H}\mathbb{I}[R>0],
\\
c_i^{(8)}
&=
\tanh
\left(
\frac{
\max(0,f_i-f_{best})
}{
\max(\operatorname{MAD}(\mathbf{f}),10^{-12})
}
\right)
\end{align}

Here $H$ denotes the recent reward history. The true optimum of a benchmark
function is not used in the context or search controller.

\begin{algorithm}[t]
\caption{Cognitive Memory-Driven Optimization (CMDO)}
\label{alg:cmdo}
\begin{algorithmic}[1]
\Require Objective $f$, bounds $(\mathbf{l},\mathbf{u})$, dimension $D$,
evaluation budget $B$, population size $N$
\Ensure Best solution $\mathbf{x}_{best}$

\State Initialize $N$ solutions uniformly in $[0,1]^D$
\State Evaluate the initial population and count all evaluations
\State Initialize working memory $\mathcal{M}_W$, episodic memory
$\mathcal{M}_E$, and consolidated memory $\mathcal{M}_C$
\State Initialize reward history and stagnation counter

\While{$n_{\mathrm{eval}} < B$}
    \State Select focal index $i$ cyclically
    \State Compute search context $\mathbf{c}_i$
    \State Retrieve contextually relevant records from
    $\mathcal{M}_W \cup \mathcal{M}_E \cup \mathcal{M}_C$

    \If{no usable recalled recipe exists \textbf{or} fresh exploration is selected}
        \State Generate a fresh recipe $\mathbf{r}=(o,s,m,g)$
        \State Set recalled source to $\varnothing$
    \Else
        \State Select a recalled recipe according to memory support
        \State Record its source experience
    \EndIf

    \State Construct search direction $\mathbf{d}$ from the current population
    \State Apply full-space or subspace geometry according to $g$
    \State Generate candidate
    $\mathbf{z}'=\operatorname{Reflect}(\mathbf{z}_i+s\mathbf{d})$
    \State Map $\mathbf{z}'$ from $[0,1]^D$ to the physical search space

    \State Compute pre-evaluation fitness scale
    $\sigma_f=\max(\operatorname{MAD}(\mathbf{f}),10^{-12})$
    \State Evaluate $f(\mathbf{x}')$
    \State Compute
    $R=\tanh((f_i-f')/\sigma_f)$

    \If{$f' < f_i$}
        \State Replace the focal solution with the candidate
    \EndIf
    \State Update the global best and stagnation state

    \If{a recalled source was executed}
        \State Update reuse count and reuse return of that source only
    \EndIf

    \State Store the new experience in working and episodic memory
    \State Consolidate eligible repeatedly useful experience
    \State Apply bounded-memory retention and eviction
\EndWhile

\State \Return $\mathbf{x}_{best}$
\end{algorithmic}
\end{algorithm}

\subsection{Fresh recipes and search geometry}

A recipe is
\begin{equation}
    \mathbf{r}=(o,s,m,g),
\end{equation}
where $o$ selects exploration, directed search, or local refinement; $s$ is the
step scale; $m$ controls the directional mixture; and $g$ specifies full-space
or subspace geometry.

Fresh recipes use
\begin{equation}
P(o=\mathrm{exploration})=0.5,\qquad
P(o=\mathrm{directed})=0.3,\qquad
P(o=\mathrm{local})=0.2,
\end{equation}
with
\begin{equation}
s\sim\operatorname{LogUniform}(0.25,2),
\qquad
m\sim\mathcal{U}(0.2,0.8)
\end{equation}

For two sampled population members,
\begin{equation}
    \boldsymbol{\delta}=\mathbf{z}_a-\mathbf{z}_b,
\end{equation}
and the stochastic component is
\begin{equation}
    \boldsymbol{\epsilon}
    =
    \boldsymbol{\eta}
    \odot
    \max(\boldsymbol{\sigma}_Z,0.01),
    \qquad
    \boldsymbol{\eta}\sim\mathcal{N}(\mathbf{0},\mathbf{I})
\end{equation}

The three directions are
\begin{align}
\mathbf{d}_{exp}
&=
m\boldsymbol{\delta}+(1-m)\boldsymbol{\epsilon},
\\
\mathbf{d}_{dir}
&=
m(\mathbf{z}_{best}-\mathbf{z}_i)+(1-m)\boldsymbol{\delta},
\\
\mathbf{d}_{loc}
&=
0.1\left[
m(\mathbf{z}_{best}-\mathbf{z}_i)+(1-m)\boldsymbol{\epsilon}
\right]
\end{align}

For subspace geometry and $D>1$, CMDO modifies
\begin{equation}
    \left\lceil\sqrt{D}\right\rceil
\end{equation}
coordinates sampled uniformly without replacement. Coordinates outside the
selected subset remain equal to those of the focal solution.

\begin{table}[t]
\centering
\small
\renewcommand{\arraystretch}{1.25}
\setlength{\tabcolsep}{5pt}

\newcolumntype{Y}{>{\centering\arraybackslash}X}

\caption{Numerical benchmark configuration.}
\label{tab:appendix_benchmarks}
\vspace{4pt}
\begin{tabularx}{\linewidth}{|l|Y|Y|Y|Y|}
\hline
\textbf{Suite} & \textbf{Functions} & \textbf{$D$} & \textbf{Budget} & \textbf{Seeds} \\ \hline
BBOB/COCO & F1, F8, F15 &  5 &  500 & 15--17 \\ \hline
CEC2017   & F1, F4, F10 & 10 & 1000 & 15--17 \\ \hline
\end{tabularx}
\end{table}

\subsection{Retrieval and negative evidence}

Records from the three memory stores are first deduplicated by experience
identifier. Similarity between the current context $\mathbf{c}$ and stored
context $\mathbf{c}_j$ is
\begin{equation}
S_j
=
\exp
\left(
-
\frac{
\operatorname{MSE}(\mathbf{c},\mathbf{c}_j)
}{
2h^2
}
\right),
\end{equation}
and the age-adjusted retrieval weight is
\begin{equation}
    W_j=S_j\rho^{a_j}
\end{equation}

Records below the relevance threshold are removed, after which the top-$k$
records by retrieval weight are considered. The current value of a record is
\begin{equation}
V_j
=
\frac{
R_j+R^{\mathrm{reuse}}_j
}{
1+n^{\mathrm{reuse}}_j
}
\end{equation}

Only positive-valued records can directly propose a recalled recipe. Retrieved
negative records using the same operator and geometry contribute a penalty
\begin{equation}
P_j
=
\sum_{q\in\mathcal{N}_j}
W_q |V_q|
\exp
\left(
-
\left|
\log\frac{s_q}{s_j}
\right|
-
|m_q-m_j|
\right),
\end{equation}
giving the final support
\begin{equation}
    Q_j=W_jV_j-\lambda P_j.
\end{equation}

Recipes with $Q_j>0$ are sampled proportionally to their support. If none
remain, a fresh recipe is used.

\begin{figure}[t]
    \centering
    \includegraphics[width=0.32\linewidth]{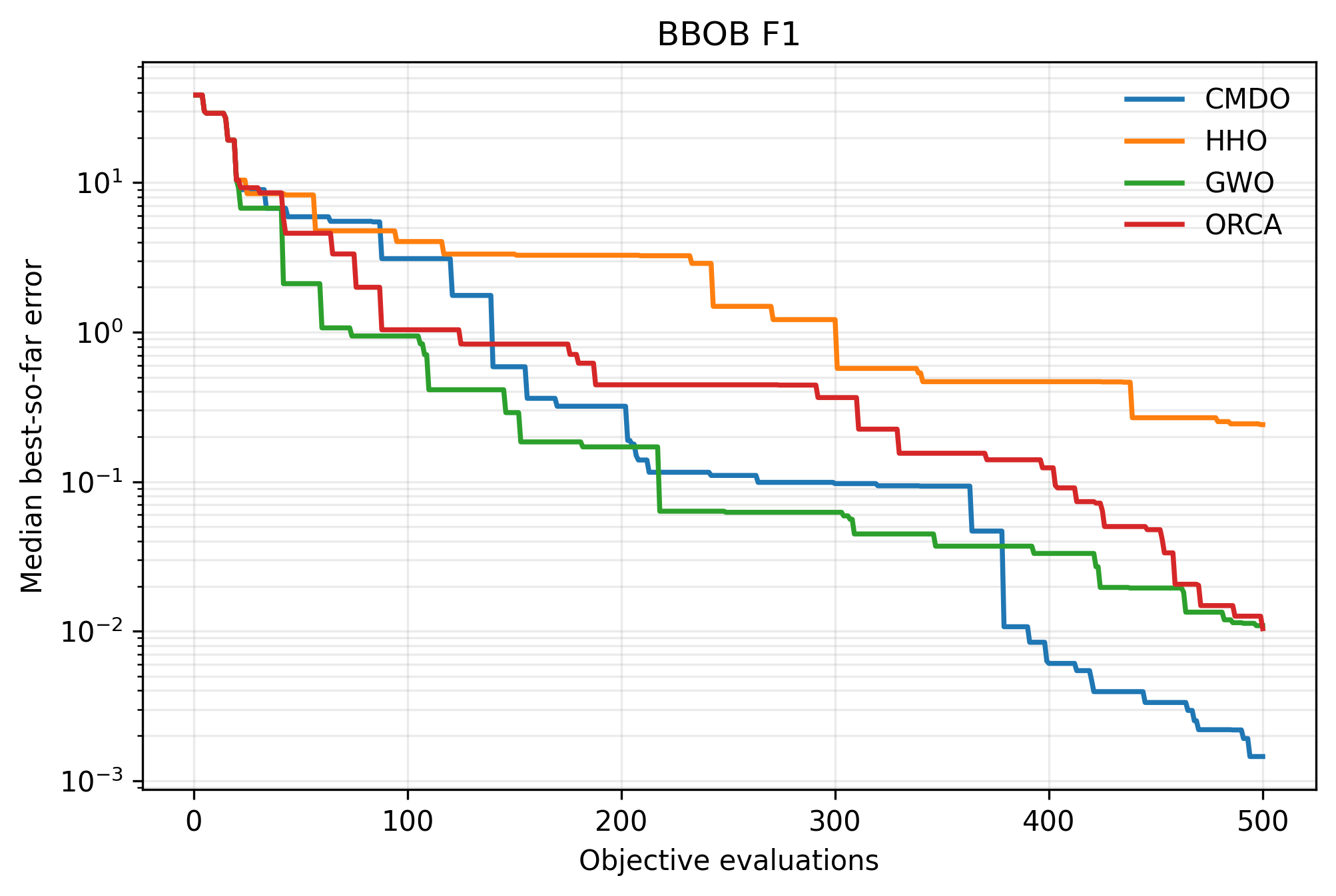}
    \includegraphics[width=0.32\linewidth]{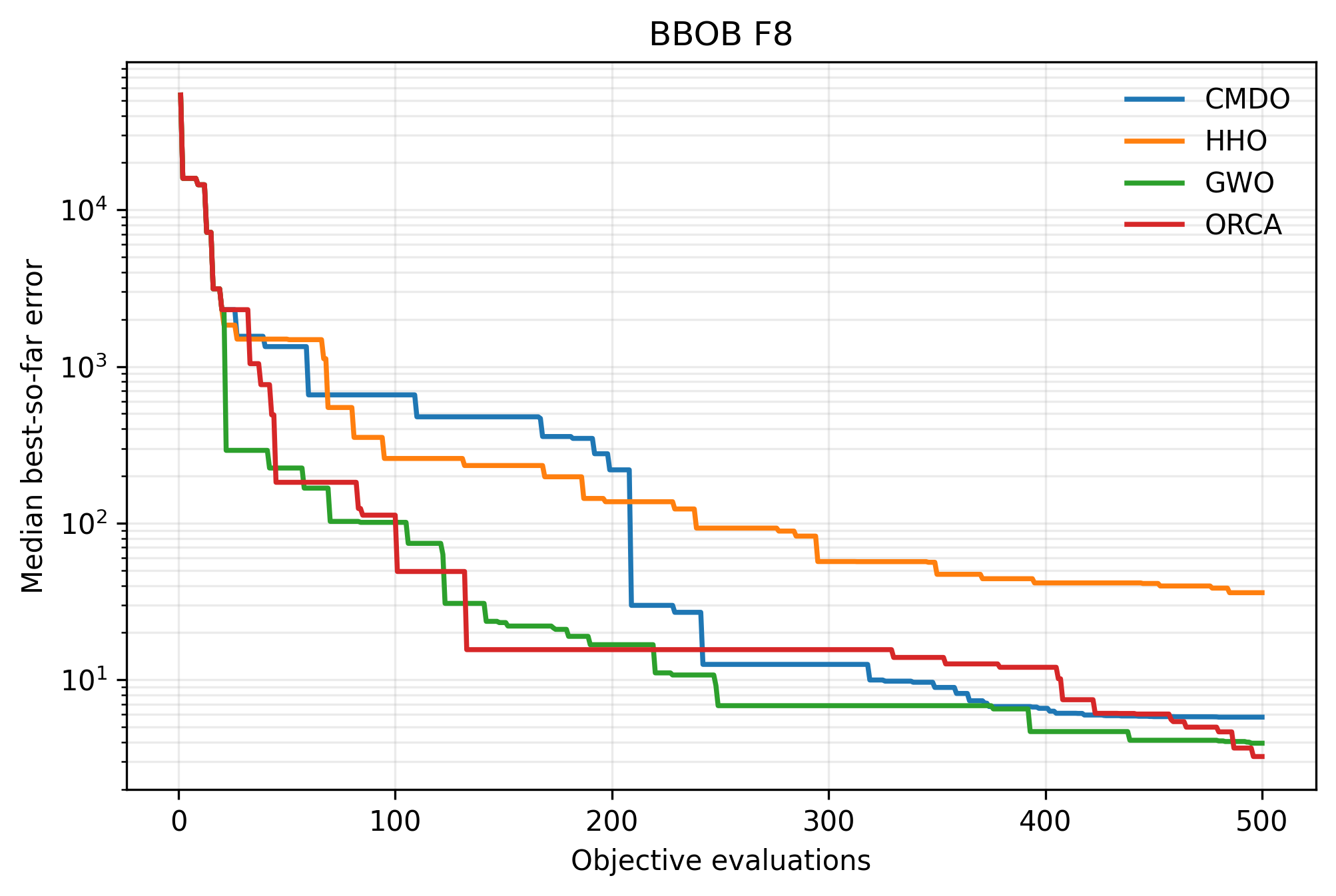}
    \includegraphics[width=0.32\linewidth]{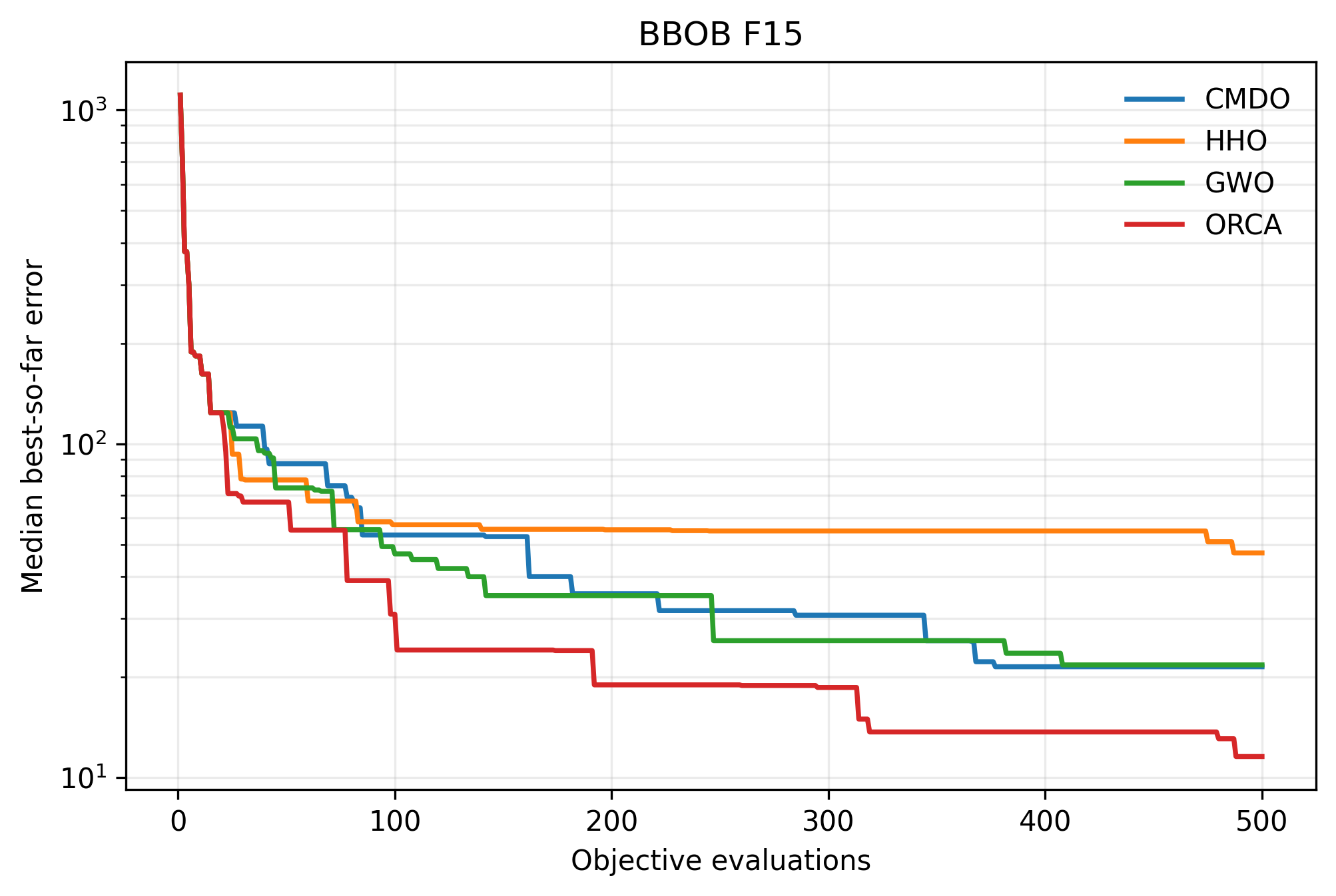}

    \vspace{0.5em}

    \includegraphics[width=0.32\linewidth]{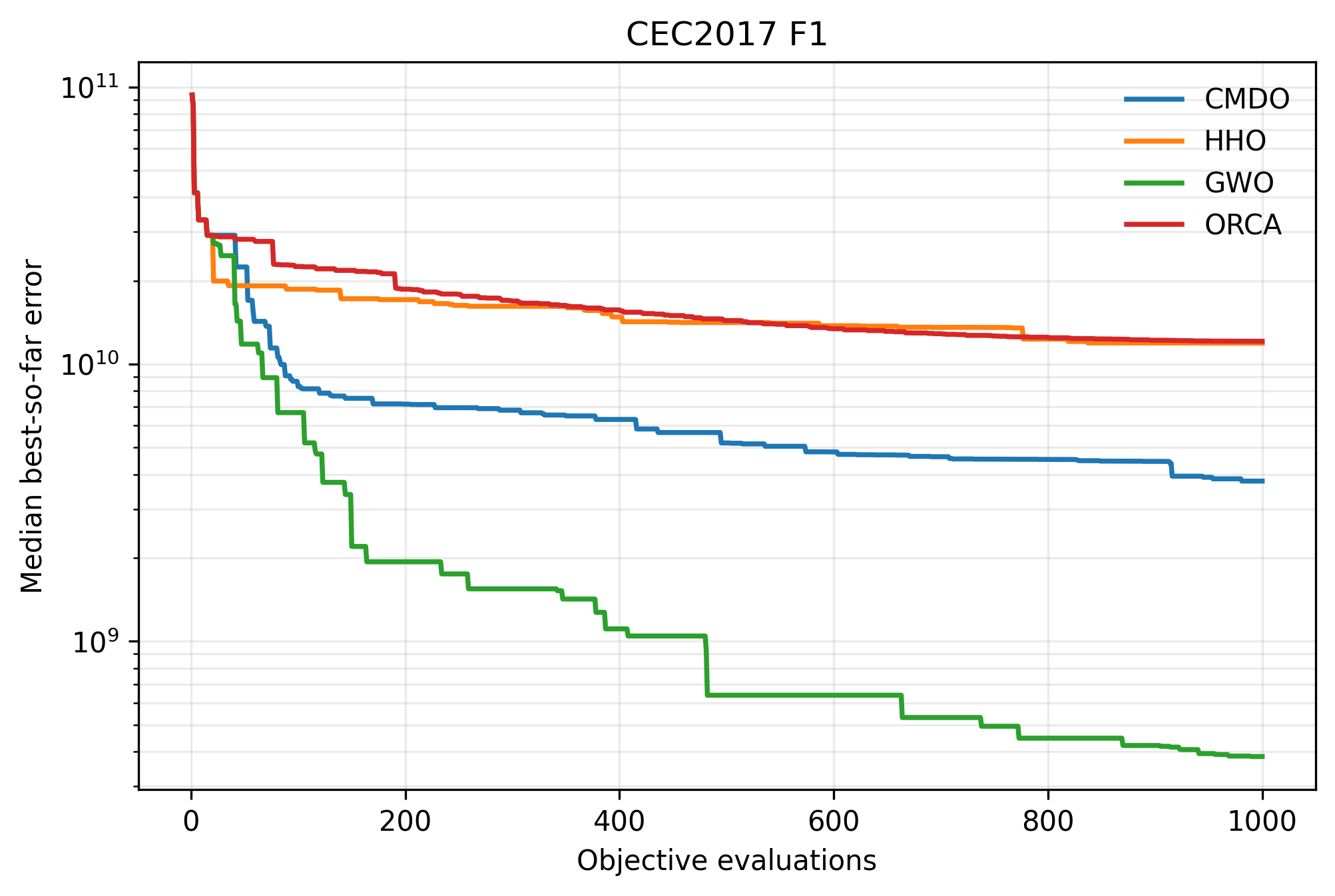}
    \includegraphics[width=0.32\linewidth]{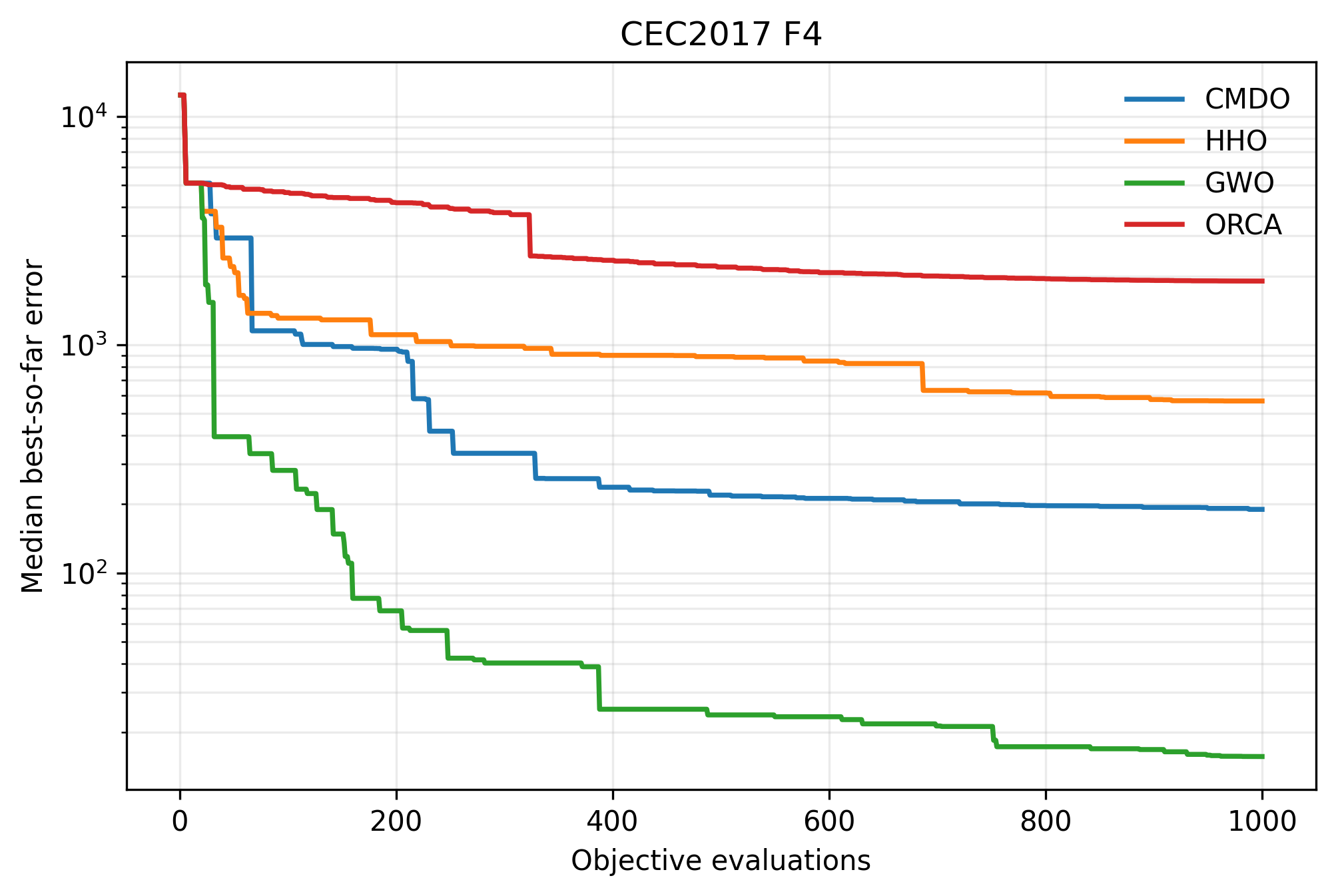}
    \includegraphics[width=0.32\linewidth]{cec2017_f10_convergence.png}

    \caption{
    Median best-so-far optimization error over seeds 15--17.
    The top row shows BBOB F1, F8, and F15; the bottom row shows
    CEC2017 F1, F4, and F10. The vertical axis is logarithmic.
    }
    \label{fig:appendix_convergence}
\end{figure}

\subsection{Memory update and consolidation}

Each completed search step generates a new experience containing the search
context, executed recipe, reward, search step, and realized displacement,
together with metadata used for later reuse. The new event is written to both
working and episodic memory, regardless of whether its reward is positive or
negative.

When a remembered recipe is actually executed, reuse feedback is assigned only
to the selected source record. An original experience becomes eligible for
consolidation after at least three selected reuses with a positive mean reuse
return.

Two records are compatible for consolidation when they have the same operator
and geometry and satisfy
\begin{equation}
\operatorname{MSE}(\mathbf{c}_p,\mathbf{c}_j)<0.01,
\end{equation}
\begin{equation}
\left|\log\frac{s_p}{s_j}\right|<0.25,
\qquad
|m_p-m_j|<0.1
\end{equation}

Compatible records update a consolidated prototype; otherwise, a new prototype
is formed. Working memory retains the newest 24 experiences, episodic memory
is bounded at 256 records using age-decayed utility for eviction, and
consolidated memory contains at most 24 prototypes.

\section{Experimental Implementation Details}
\label{app:implementation}

\subsection{Benchmark protocol}

The numerical experiments use BBOB/COCO functions F1, F8, and F15 at
$D=5$ with 500 evaluations, and CEC2017 functions F1, F4, and F10 at
$D=10$ with 1000 evaluations. Instance 1 and seeds 15--17 are used for the
reported benchmark panel. All objective evaluations, including initialization,
count toward the stated budget.

\section{Additional Convergence Results}
\label{app:convergence}

The main paper presents CEC2017 F10 as a representative convergence example.
Figure~\ref{fig:appendix_convergence} shows the remaining recorded convergence
trajectories together with F10 for completeness. Curves report the median
best-so-far error across seeds 15--17.

These trajectories complement the final-error tables by showing when
improvements occur within the fixed evaluation budget. They also illustrate
the problem-dependent evolution of CMDO's search behavior across the selected
landscapes.

\section{Photovoltaic Model and Experimental Details}
\label{app:pv}

\subsection{Two-diode photovoltaic model}

The photovoltaic experiment estimates the seven parameters
\begin{equation}
\boldsymbol{\theta}
=
[
I_{\mathrm{ph}},
I_{01},
I_{02},
R_s,
R_p,
a_1,
a_2
]
\end{equation}
of the two-diode model. For measured terminal voltage $V$ and current $I$, the
model is defined implicitly as
\begin{equation}
I
=
I_{\mathrm{ph}}
-
I_{01}
\left[
\exp\left(
\frac{V+IR_s}{a_1V_T}
\right)-1
\right]
-
I_{02}
\left[
\exp\left(
\frac{V+IR_s}{a_2V_T}
\right)-1
\right]
-
\frac{V+IR_s}{R_p},
\label{eq:appendix_twodiode}
\end{equation}
where
\begin{equation}
    V_T=\frac{N_s kT}{q}
\end{equation}

Here $k$ is the Boltzmann constant, $q$ is the elementary charge, $T$ is the
cell temperature in Kelvin, and $N_s=1$ for the RTC France cell.

The optimization objective is
\begin{equation}
\mathrm{MAE}(\boldsymbol{\theta})
=
\frac{1}{26}
\sum_{j=1}^{26}
\left|
I_j^{\mathrm{meas}}
-
I_j^{\mathrm{model}}(\boldsymbol{\theta})
\right|
\end{equation}

RMSE is calculated after optimization as an additional fit-quality measure.

\subsection{Parameter bounds and representation}

\begin{table}[t]
\centering
\small
\renewcommand{\arraystretch}{1.25}
\setlength{\tabcolsep}{6pt}

\newcolumntype{Y}{>{\centering\arraybackslash}X}

\caption{Parameter bounds for the RTC France two-diode model.}
\label{tab:pv_bounds_appendix}
\vspace{4pt}
\begin{tabularx}{\linewidth}{|l|Y|Y|}
\hline
\textbf{Parameter} & \textbf{Lower bound} & \textbf{Upper bound} \\ \hline
$I_{\mathrm{ph}}$ (A) & 0          & 1         \\ \hline
$I_{01}$ (A)          & $10^{-15}$ & $10^{-3}$ \\ \hline
$I_{02}$ (A)          & $10^{-15}$ & $10^{-3}$ \\ \hline
$R_s$ ($\Omega$)      & 0          & 0.5       \\ \hline
$R_p$ ($\Omega$)      & 0.001      & 100       \\ \hline
$a_1$                 & 0.5        & 5         \\ \hline
$a_2$                 & 1          & 5         \\ \hline
\end{tabularx}
\end{table}

The search is performed in a common normalized seven-dimensional space.
The parameter bounds used for the RTC France two-diode model are summarized
in Table~\ref{tab:pv_bounds_appendix}. $I_{01}$ and $I_{02}$ are decoded
logarithmically because their ranges span
multiple orders of magnitude; the remaining parameters are mapped linearly.

The experiment uses the 26 measured RTC France $I$--$V$ observations at
$33^{\circ}\mathrm{C}$ and $1000~\mathrm{W/m^2}$ reported by
\citet{muhammad2019pv}. Parameter bounds follow the RTC France two-diode
settings used by \citet{qin2024pv}. Each optimizer receives 2000 objective
evaluations for seeds 101--110.

For every candidate parameter vector, the implicit model equation is solved at
all measured voltages before MAE is computed. The same model solver, measured
data, parameter encoding, and objective function are used for all compared
optimization methods.

\end{document}